\documentclass[runningheads]{llncs}
\usepackage[T1]{fontenc}
\usepackage{caption}
\usepackage{xcolor}
\usepackage{graphicx}
\begin{document}
\title{DEF-YOLO: Leveraging YOLO for Concealed Weapon
Detection in Thermal Imaging}
%

\author{Divya Bhardwaj\textsuperscript{*}\and
Arnav Ramamoorthy \and
Poonam Goyal\textsuperscript{*}  }
\authorrunning{Divya Bhardwaj et al.}
%
\institute{Birla Institute of Technology \& Science, Pilani Campus, Pilani, Rajasthan, India\and
\email{\{p20180013, f20220007, poonam\}@pilani.bits-pilani.ac.in}\\
}
\maketitle              
\begin{abstract}
 Concealed weapon detection aims at detecting weapons hidden beneath a person's clothing or luggage. Various imaging modalities like Millimeter Wave, Microwave, Terahertz, Infrared, etc., are exploited for the concealed weapon detection task. These imaging modalities have their own limitations, such as poor resolution in microwave imaging, privacy concerns in millimeter wave imaging, etc. To provide a real-time, 24 $\times$ 7 surveillance, low-cost, and privacy-preserved solution, we opted for thermal imaging in spite of the lack of availability of a benchmark dataset. We propose a novel approach and a dataset for concealed weapon detection in thermal imagery. Our YOLO-based architecture, DEF-YOLO, is built with key enhancements in YOLOv8 tailored to the unique challenges of concealed weapon detection in thermal vision. We adopt deformable convolutions at the SPPF layer to exploit multi-scale features; backbone and neck layers to extract low, mid, and high-level features, enabling DEF-YOLO to adaptively focus on localization around the objects in thermal homogeneous regions, without sacrificing much of the speed and throughput. In addition to these simple yet effective key architectural changes, we introduce a new, large-scale Thermal Imaging Concealed Weapon dataset, TICW, featuring a diverse set of concealed weapons and capturing a wide range of scenarios. To the best of our knowledge, this is the first large-scale contributed dataset for this task. We also incorporate focal loss to address the significant class imbalance inherent in the concealed weapon detection task. The efficacy of the proposed work establishes a new benchmark through extensive experimentation for concealed weapon detection in thermal imagery.

\end{abstract}
\section{Introduction}
Efforts have been afoot towards securing one's life in public places. e.g., airports, historical places, etc., where advanced automated systems have been installed to detect anomalous substances, such as weapons, while scanning passengers and their baggage. In general, these scanning machines utilize electromagnetic radiation to penetrate the items, thereby creating an inherent view of their content. X-ray-based scanners are used only for scanning baggage, as they are prone to a carcinogenic effect on humans. Millimeter-wave (MMW) imaging scanners are considered to be better compared to X-ray-based scanners for humans to detect concealed weapons; however, they are prone to violating privacy concerns. Moreover, MMW images have severe noise interference, and the size of the concealed object is also very small \cite{li2025lightweight}.  Terahertz (THz) based systems have been proven to be a better alternative to X-ray and MMW systems. However, their higher installation cost and lower imaging resolution adversely impact their feasibility in real-world deployment in public places. Hence, a common practice that can be seen worldwide across all airports is to scan the baggage for weapons using X-ray scanners, and humans are made to first pass through a metal detector, followed by security personnel who verify the presence of weapons or anomalous substances manually using hand-held detectors. This practice is not only risk-prone but also incurs lower throughput.

Thermal imaging is a low-cost, illumination-invariant alternative to the aforementioned systems, which not only ensures privacy but is also installation-friendly. It operates based on the principle of heat emission, i.e., objects are detected based on the heat they emit. The clarity of an object’s appearance in the image improves with a greater temperature difference between the concealed object and the human body. To enhance detection, individuals can be asked to pass through a temperature-controlled environment, such as those commonly found in airports. 
In this setting, concealed metallic weapons cool down much faster compared to the human body due to their thermal properties, creating a detectable temperature contrast. Under such circumstances, it is possible to detect weapons using thermal imaging systems. The main aim of this paper is to present a real-time learning-based framework that can detect concealed weapons from thermal images of humans, without exposing privacy and radiation threats.

The majority of the existing work for weapon detection has utilized MMW and THz-based imaging \cite{su2024enhancing,cheng2025enhancing,cheng2022improved,yang2022transformer}. Whereas thermal imaging has been relatively less explored. Thermal imaging has been used mainly in fusion with visible images to carry out CWD task \cite{hussein2016alternative,hussein2017multisensor,raturi2019adocw,bhavana2022infrared,goyal2021infrared,gosain2021concealed}. Existing methods, such as \cite{hussein2016alternative,hussein2017multisensor,bhavana2022infrared,gosain2021concealed} are based on traditional computer vision algorithms, e.g., Discrete Wavelet Transform (DWT), dimensionality reduction, and low-rank representation, to detect weapons based on high-frequency details and concise feature representations, but lack in generalization. These methods use RGB deep learning models with fine-tuning on the CWD data without any modification in the model architecture. For example, Faster R-CNN is used as it is for the CWD task after fusing the thermal and visible images \cite{raturi2019adocw}; Veranyurt et al. evaluated the performance of their own custom dataset for concealed pistol detection on various deep learning models, such as SSD, YOLOv2, Tiny-YOLO, Mask R-CNN, etc. This dataset has 600 thermal images, out of which only 380 images have the 11 subjects carrying a pistol, mostly with a front and back view \cite{veranyurt2023concealed}. However, it is not publicly available. We constructed our TICW dataset with 6000 images, where 22 subjects are carrying multiple weapons with different views, different positions, and wearing different clothing. Our dataset bridges the gap in the thermal domain for the CWD task, which is crucial to achieve a near-real-time and viable solution for public surveillance. 

The presence of a concealed weapon in thermal images depends on its heat emission. Moreover, the temperature gradient between the human body and the weapon plays a significant role in weapon visibility in thermal imagery. Hence, we propose a YOLOv8-based architecture that adaptively learns about the concealed weapon using deformable convolution and can be deployed for real-time surveillance applications. 
We summarize our contributions as follows:

\begin{itemize}
\item We modify the YOLOv8 architecture for CWD on thermal images using deformable convolution in SPPF and a few layers of C2f of YOLOv8, which adaptively learns the location of concealed weapons. Also, we integrated focal loss, which prevents the network from being biased towards easy and majority samples. 
\item We constructed our own concealed weapon detection dataset, TICW, using thermal modality. To the best of our knowledge, this is the largest thermal dataset having 6k images for the CWD task. The dataset is prepared for multiple weapons at various positions using different postures, making it diverse, hence more suitable for real-time surveillance applications.
\end{itemize} 

\section{Related Work}
The CWD can be carried out using various imaging modalities, such as Microwave, MMW, THz, X-rays, Infrared, Thermal, etc. CWD methods can be categorized into two categories: 1) multi modality: these methods mainly used thermal/infrared image with corresponding visible image for CWD task \cite{bhavana2022infrared,hussein2016alternative,hussein2017multisensor,goyal2021infrared}; 2) single modality:  these methods use either MMW imaging \cite{wang2021self,yang2022transformer,cheng2022improved} or THz imaging \cite{cheng2024few,cheng2025enhancing,su2024enhancing}. But very few methods work on only infrared/thermal imaging \cite{khor2024automated,veranyurt2023concealed}.

\textbf{Multi-modality methods.}
 Bhavana et al. \cite{bhavana2022infrared} used a Latent low-rank method to fuse the infrared and visible images for finding the concealed object beneath a person's clothing. On the other hand, \cite{hussein2016alternative,hussein2017multisensor,gosain2021concealed,goyal2021infrared,raturi2019adocw} fused visible and infrared modality images using the DWT. Hussein et al. \cite{hussein2016alternative} used DWT for fusion of infrared and visible images, followed by segmentation using thresholding. While \cite{hussein2017multisensor}, uses DWT with hybrid dimensionality reduction block to fuse thermal and visual images. Then K-Means is used for detecting threats, followed by classification using a support vector machine. The comparison between DWT, Discrete Cosine Transform, and guided filter algorithm is presented \cite{goyal2021infrared} for fusion of infrared and visible images, stating DWT outperforms the other two techniques with low noise and high fusion rate. The method in \cite{raturi2019adocw} preprocesses the infrared and visible images using the Canny edge detector and then applies non-maximum suppression to reduce the false detections; lastly, trains the Faster R-CNN for detecting concealed weapons.

\textbf{Single modality methods.}
In order to detect small objects from MMW images, \cite{wang2021self} proposed an attention fusion network that exploits multi-scale features from ResNet. They showed the performance of their approach on two datasets, Active MMW and Passive MMW. Yang et al. \cite{yang2022transformer} used a hierarchical transformer-based backbone with an attention module for detecting concealed objects in passive MMW images. The Single Shot MultiBox Detector was improved to detect concealed objects from THz images. The authors made the modification by introducing a ResNet, feature fusion module, and an attention mechanism \cite{cheng2022improved}. Cheng et al. \cite{cheng2024few} introduced a novel pseudo-annotation method tailored for few-shot object detection in sub-THz images to overcome labelled data and class imbalance issues. Su et al. \cite{su2024enhancing} modified YOLOv8 by replacing some layers with wavelet convolution and incorporating a wavelet attention module for detecting concealed objects from Active MMW images. The recent work proposed the Adaptation-YOLO, a framework that is based on YOLOv8. They proposed two major components: an adaptive context-aware attention network and a dynamic adaptive convolution block to detect concealed objects in THz images \cite{cheng2025enhancing}. The authors used VGG-16 for classifying whether an image contains a pistol or not and YOLOv3 for detecting the concealed pistol \cite{veranyurt2023concealed}.
Using pre-processing techniques like Fuzzy C-means clustering, Region-of-Interest cropped images enhanced the performance of ResNet-50 for detecting concealed objects in thermal images \cite{khor2024automated}. The most recent work for detecting concealed handgun from thermal imaging used YOLOv3 \cite{munoz2025concealed}. 

\textbf{Datasets on CWD task.} Researches have been carried out in constructing the dataset for various modalities to perform the CWD task.  1) \textit{THz Imaging Dataset.} The Active THz Dataset contains a total of 3,157 images, out of which only 1,194 images have concealed objects. There is a total of 11 categories of concealed objects, i.e., gun, kitchen knife, cell phone, ceramic knife, metal dagger, water bottle, key chain, cigarette lighter, leather wallet, scissors, and unknown \cite{liang2021active}. 2) \textit{MMW Imaging Dataset.} The BHU-1024 dataset has 1921 passive MMW images with a size of 160$\times$80. It comprises 4 classes: ceramic knife, metallic knife, mobile phone, and simulated gun. Objects are concealed in the human body on the back, waist, chest, legs, etc \cite{yang2022transformer}. The AMMW-HiSC captures 36,880 active MMW images with a resolution of 5 millimeters and 31 categories of concealed objects; lipsticks, grenades, handguns, baby creams, lighters, etc, to name a few. The objects in this dataset are less than 32 pixels in size \cite{wang2021self}. Another passive MMW imaging dataset contains total of 3309 images where 2846 images have 12 different concealed objects with a resolution of 125$\times$195. A cutter, a clay, a simulated gun, sugar, frozen peas, a bag with metal pieces, flour, a water bottle, and a hydrogen peroxide bottle were used for concealing on human body parts, chest, forearm, thigh, etc, \cite{lopez2018using}. 3) \textit{X-ray Imaging Dataset.} The Si-Xray dataset contains 1,059,231 X-ray images, out of which 8929 have prohibited items such as hammers, scissors, and pliers. These weapons are kept inside the baggage. This is the largest publicly available dataset for the CWD task \cite{Miao2019SIXray}. 4) \textit{Thermal Imaging Dataset.} A concealed pistol detection was constructed by \cite{veranyurt2023concealed} using thermal imaging, which has 600 images out of which 380 images have the concealed pistol belonging to 11 subjects. Another dataset contains 1100 thermal images with 562 containing the concealed object \cite{khor2024automated}.
The authors Raturi et al. \cite{raturi2019adocw} created their own custom dataset with a training set of 9084 samples containing images with and without weapons; a testing set of 1000 images containing 650 images with weapons. None of the thermal datasets mentioned are publicly available, nor are they made available on demand.    

\section{Thermal Imaging Concealed Weapons Dataset (TICW)}
One of the major contributions of our work is creating the thermal imaging concealed weapon (TICW) dataset. Considering the real-time surveillance, we planned to capture images in many possible scenarios for robust training. This is the biggest dataset for the CWD task in thermal modality.
\begin{table}[t]
\centering
\resizebox{0.6\linewidth}{!}{
\begin{tabular}{|c|cc|cc|cc|}
\hline
         & \multicolumn{2}{c|}{Training Set}       & \multicolumn{2}{c|}{Validation Set}     & \multicolumn{2}{c|}{Testing Set}        \\ \hline
Class    & \multicolumn{1}{c|}{Images} & Instances & \multicolumn{1}{c|}{Images} & Instances & \multicolumn{1}{c|}{Images} & Instances \\ \hline
All      & \multicolumn{1}{c|}{4800}   & 7941      & \multicolumn{1}{c|}{600}    & 998       & \multicolumn{1}{c|}{600}    & 1000      \\ \hline
Cleaver  & \multicolumn{1}{c|}{1236}   & 1274      & \multicolumn{1}{c|}{147}    & 152       & \multicolumn{1}{c|}{171}    & 177       \\ \hline
Gun      & \multicolumn{1}{c|}{1984}   & 2184      & \multicolumn{1}{c|}{277}    & 302       & \multicolumn{1}{c|}{239}    & 260       \\ \hline
Knife    & \multicolumn{1}{c|}{2740}   & 3730      & \multicolumn{1}{c|}{327}    & 459       & \multicolumn{1}{c|}{356}    & 478       \\ \hline
Scissors & \multicolumn{1}{c|}{752}    & 753       & \multicolumn{1}{c|}{84}     & 85        & \multicolumn{1}{c|}{84}     & 85        \\ \hline
\end{tabular}}
\vspace{0.2cm}
\captionsetup{font=scriptsize}
\caption{Classwise distribution of TICW dataset for training, validation, and testing.}
\label{tab:ticw data}
\end{table}

\begin{figure}[t]
\begin{center}
\includegraphics[width=0.7\linewidth]{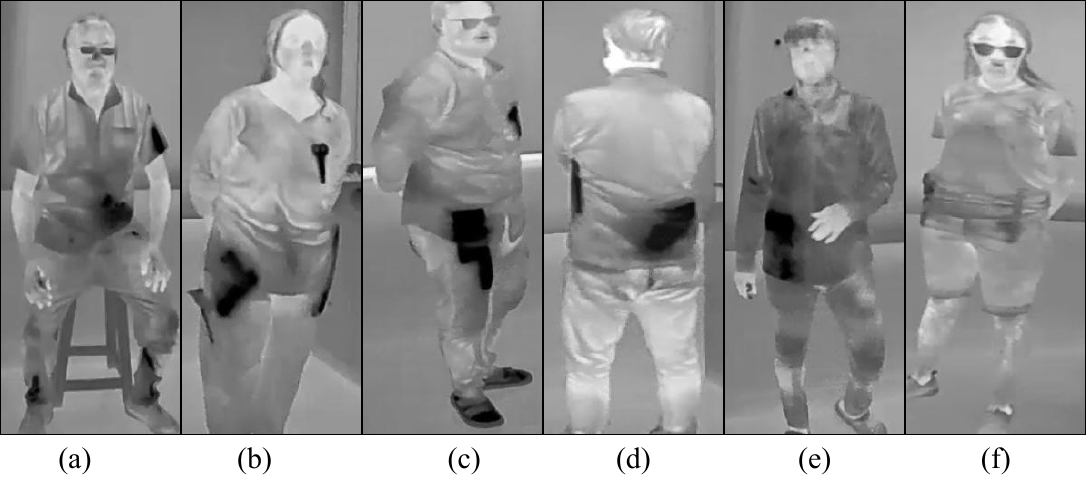}
\end{center}
\vspace{-0.5cm}
\captionsetup{font=scriptsize}
\caption{Subject in (a) sitting position carrying scissors in right leg, knife in left leg, gun in waist, and knife in arm. (b) standing position with front viewpoint carrying a knife in the right thigh, a gun and a knife in the waist, and scissors in the chest. (c) side viewpoint carrying a cleaver at the waist and scissors in the chest. (d) back viewpoint carrying a gun and a cleaver. (e) front viewpoint carrying a gun in a thin jacket. (f) front viewpoint and less temperature gradient between the weapon, gun, knife concealed in the waist, and the subject's body.}
\label{fig:ticw}
\end{figure}

\textbf{Data Capture.}
We constructed the TICW dataset using an Axis Q1942-E thermal network camera. The thermal camera has a resolution of 640$\times$480 with a frame rate of 9 frames per second.  
The dataset is captured with the help of 16 male and 9 female subjects wearing different clothing, for example, a shirt, a thin jacket, jeans, shorts, pants, cotton clothes, etc. The subjects were instructed to use different positions to conceal weapons in various poses, such as standing, sitting, etc., and the images of the subjects were captured from different viewpoints: front, back, and side. We have used different kinds of knives, guns, cleavers, and scissors. The 25 subjects aged between 22-40 used these weapons to hide inside their clothing at one or multiple locations such as the waist, chest, back, thigh, legs, hands, abdomen, etc. These images are captured in varied temperatures, and the subjects were instructed to conceal weapons for varying durations so that there is a variation in the visibility of weapons in the images. To the best of our knowledge, this is the largest, comprehensive CWD dataset in thermal imaging with a total of 6k images. Our dataset captures diverse scenarios required for detecting concealed weapons and is hence suitable for real-time deployment for public surveillance. We present the class distribution of our dataset in the Table. \ref{tab:ticw data}.

\textbf{Annotations.}
Firstly, we cropped the person ROIs from the thermal images, and then from the ROIs, the weapons were annotated. We used the Roboflow tool \cite{roboflow} for creating bounding boxes on the weapons and classifying them. There are four classes: cleaver, knife, gun, and scissors. We assigned 5 human annotators to mark the bounding boxes and crop the person ROIs from thermal images. The annotators were instructed to mark the bounding box as tight as possible. The TICW dataset annotations are available in MS-COCO (JSON files), Pascal VOC (XML files), and YOLO format (TXT files). We show the samples of our dataset in Figure \ref{fig:ticw}. Our dataset will be made publicly available on the corresponding author's GitHub page.

\section{Method}
Considering a real-time solution for concealed weapon detection from thermal images. We first evaluated various existing YOLO object detectors. These object detectors are trained on the MS-COCO dataset, and we fine-tune YOLOv5, YOLOv8, and YOLOv11 on the TICW dataset. Based on this analysis, YOLOv8 performs better than YOLOv5 and YOLOv11, as stated in the Table. \ref{tab:sota}. Hence, we chose YOLOv8 as our baseline model and modified it to perform specifically for CWD tasks on thermal images.

\begin{figure}[t]
     \begin{center}
         
     \includegraphics[height=9.5cm, width=0.99\linewidth]{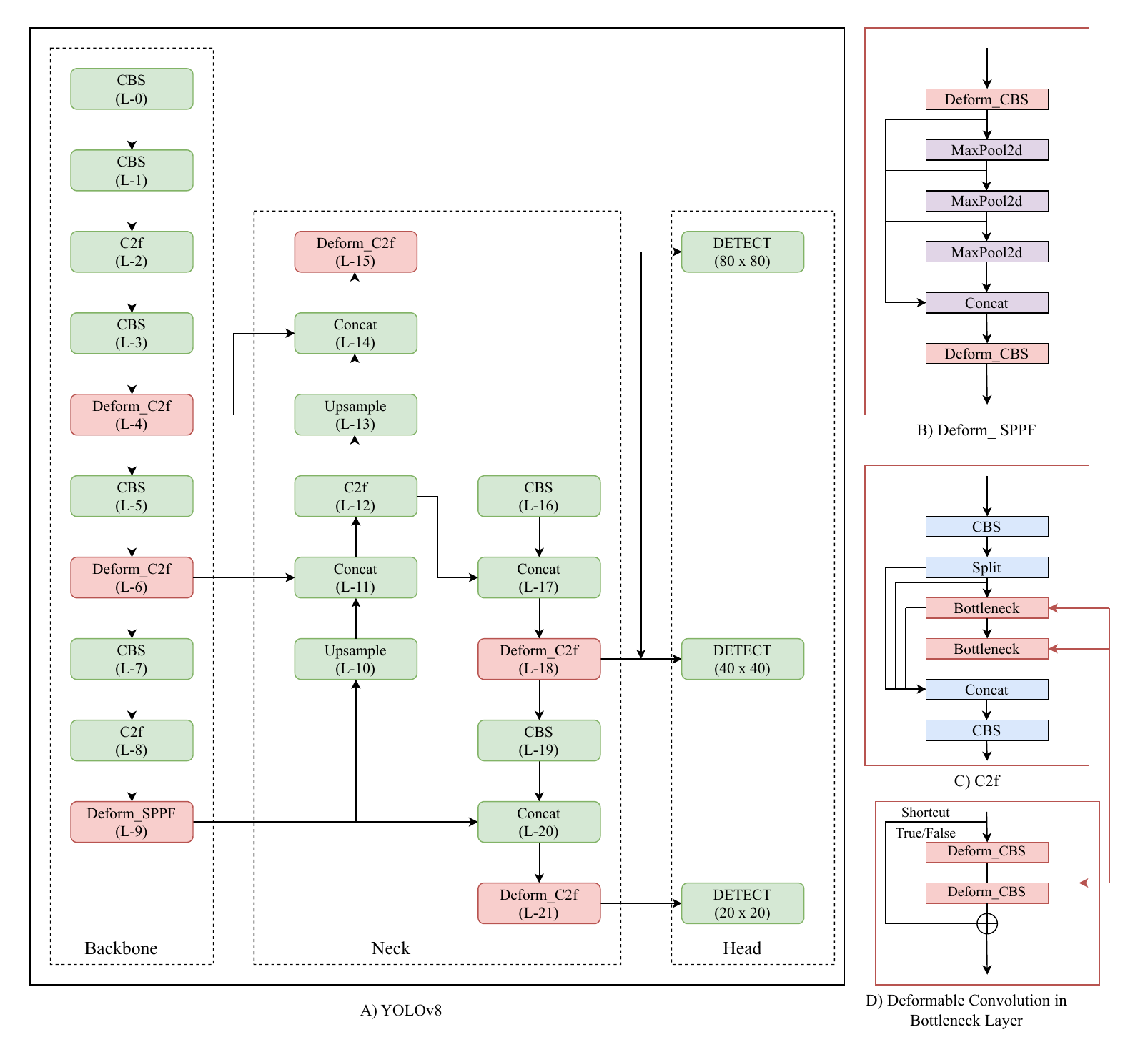}
     \end{center}
\vspace{-0.2cm}
\captionsetup{font=scriptsize}
\caption{The proposed DEF-YOLO for CWD with modified modules in red.}
\label{fig:deffYOLO}
\end{figure}

\subsection{Overview} 
We propose two modifications in the YOLOv8 architecture using deformable convolution \cite{zhu2019deformable} and called the model as DEF-YOLO (Deformable YOLO). We apply the modifications in a smaller version of YOLOv8 to demonstrate the proposed approach (see Fig. \ref{fig:deffYOLO}). We exploited multi-receptive feature from the SPPF layer to detect the concealed weapon in thermal images using deformable convolution. Next, the low, mid, and high-level features are made adaptive to learn the dynamics of concealed weapons by replacing the convolution with deformable convolution in the bottleneck block of layers 4, 6, 15, 18, and 21 of YOLOv8. We also integrated focal loss with the loss function of YOLOv8. Downweighting the effect of majority classes and upweighting the rare classes helps the network to balance the bias among all the classes.

\subsection{Deformable Convolution}
A convolution operation has a fixed, regular grid that samples the input feature map, limiting the adaptability to handle geometric transformations of objects. To overcome this limitation,  deformable convolution adds an offset to each position of a regular grid, which is learned and shifted according to the object's geometry. 

The grid $G=\{(-1,-1), (-1,0), ....., (0,1), (1,1)\}$ has the weights at each location. Given a feature $f$, with each pixel location $i_0$ and kernel offset $i_n$, which enumerates all locations of $G$. The standard convolution is represented by $out(i_0)$. 

\begin{equation}\label{eq1}
out(i_0) = {\sum_{n} w(i_n) . f(i_0 + i_n)}
\end{equation}
 whereas, deformable convolution works by integrating $\delta_n$, which makes the grid irregular. The deformable convolution $deformout(i_0)$, learns an offset $\delta_n$ by using bilinear interpolation. 
 \begin{equation}\label{eq2}
deformout(i_0) = {\sum_{n} w(i_n) . f(i_0 + i_n + \delta_n)}
\end{equation}

\begin{figure}[t]
\begin{center}
\includegraphics[height=3cm, width=0.5\linewidth]{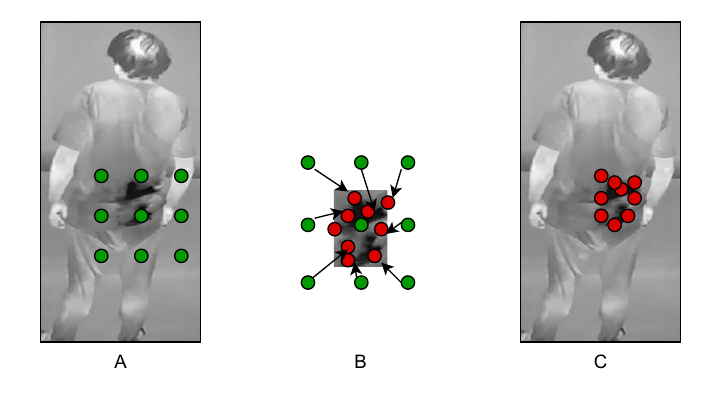}
\vspace{-0.5cm}
\captionsetup{font=scriptsize}
\caption{A) Standard convolution using a regular grid is shown in green dots. B) Deformable convolution using an irregular grid in red dots with offset direction. C) Final deformed shape showing improved localization.}
\label{fig:conv_deform}
\end{center}
\end{figure}

Fig. \ref{fig:conv_deform} shows the difference between the standard convolution and the deformable convolution. The standard convolution using a regular grid in green dots is shown, which is far from a concealed weapon visible in Fig. 3A. Fig. 3B \& C show the direction of shift towards the actual boundary of the concealed weapon and achieve the deformed shape in red dots, respectively.

\subsection{Modifications in YOLOv8}
\textbf{SPPF Layer.} The adaptability of deformable convolution to learn offsets based on geometric variation of objects, for example, size, shape, and texture, makes deformable convolution more suitable for the CWD task in thermal images. The geometry of the concealed weapon in a thermal varies depending on the heat emitted by the object. The SPPF layer of YOLOv8 is responsible for capturing features at multi-receptive fields. This layer consists of a CBS block, three maxpool layers, a concatenation operation, and lastly a CBS block. The CBS block has a configuration of a convolution layer followed by batch normalization and SiLU activation.  We replace the convolution layers with a deformable convolution layer in the CBS block of the SPPF layer. We represent the modified layer with Deform\_CBS, and the modified SPPF block is shown in Fig. \ref{fig:deffYOLO}B with the Deform\_SPPF block.

\textbf{Backbone and Neck Layers.} We replaced the convolution layers of the bottleneck in layers 4, 6, 15, 18, and 21 with deformable convolution (refer Fig.\ref{fig:deffYOLO}, where layer 4 is represented as L-4, layer 6 as L-6, and so on). We named these modified C2f blocks Deform\_C2f in the Fig. \ref{fig:deffYOLO}. Layers 4 and 6 are a part of the backbone of the YOLOv8 architecture, responsible for low-level and mid-level feature extraction. Layers 15, 18, and 21 are from the neck, responsible for high-level feature extraction. With this combination of layers, we are adapting the low, mid, and high feature layers to learn the dynamics of concealed weapons differently. Layers 4, 6, 15, 18, and 21 are C2f blocks in YOLOv8 architecture and have configuration as shown in Fig. \ref{fig:deffYOLO}C. We changed the convolution layers of the bottleneck block to deformable convolution, shown in Fig. \ref{fig:deffYOLO}D. The bottleneck block has two CBS blocks with an option of adding a residual connection.

\textbf{Strategic Placement of Modifications.} 
The SPPF layer in YOLOv8 aggregates multi-scale spatial features via max-pooling at different kernel sizes (e.g., 5×5, 9×9, 13×13). However, standard convolution layers have fixed geometric structures, which may not effectively adapt to irregular shapes or deformations, especially important in thermal images where weapons may be concealed under clothing, leading to non-rigid patterns. Replacing standard convolutions with deformable convolutions in SPPF helps adaptively capture features at varying scales, making YOLOv8 more robust to detect concealed weapons in thermal images.
The low-level feature (layer-4) in YOLOv8 helps in learning the contours and texture of the concealed object. The mid-level features (layer-6) are responsible for learning object parts, shape hints. And the high-level features (layers 15, 18, and 21) focus on learning the semantics of different classes of objects.  We used the deformable convolution in the bottleneck block to avoid vanishing gradients, which exists more for thermal images as they lack detail. The bottleneck block has residual connections between CBS blocks, which helps in retaining information; hence, modifying this layer makes the network learn more adaptively about the semantics of concealed objects in thermal images. 
With these modifications of deformable convolution in YOLOv8, our proposed framework DEF-YOLO can better adapt to the deformable, occluded, and noisy nature of concealed weapons in thermal images, leading to higher detection accuracy and robustness.

\subsection{Learning Objective}

YOLOv8 uses the loss as given below: 

\begin{equation}\label{YOLOloss}
L_Y= 7.5 * box + 0.5*cls+ 1.5 *dfl
\end{equation}

\noindent
where $box$ is the bounding box regression for improving localization accuracy; $cls$ is binary cross entropy loss used for the classification task; $dfl$ is distributed focal loss that is used for precise localization.

We used the focal loss \cite{lin2017focal} to handle the class imbalance present in our TICW dataset. As depicted in Table. \ref{tab:ticw data}, there is enough data skewness present among various classes in the TICW dataset. The focal loss typically replaces or complements the objectness or classification components. In our case, we added it to guide objectness learning more effectively on harder samples. e.g., a cleaver and scissors. The focal loss is defined as:

\begin{equation}\label{focalloss}
L_f(p_t) = -\alpha_t(1-p_t)^\gamma \log(p_t)
\end{equation}

\noindent
where $p_t=p$ if for the positive class, otherwise $p_t=1-p$; $p$ is the predicted probability for the class; $\alpha_t$ is the balancing factor for class imbalance, and $\gamma$ is the focusing parameter to reduce loss for easy examples. We use $\alpha_t$=0.25 and  $\gamma$=1.5 in our experiments. We integrate two losses $L_y$ and $L_f$ as follows:

\begin{equation}\label{totalloss}
L_T = L_Y+ 0.5*L_f
\end{equation}

The total loss incorporates bounding box regression, classification, distributed focal loss, and objectness-guided focal loss and is utilized to train the proposed DEF-YOLO.

\section{Experimental Setup}
We train our model in Pytorch for 200 epochs using the SGD optimizer with an initial learning rate of $1e^{-2}$, a warmup of 30 epochs, and a batch size of 16, on a Nvidia A100 80GB GPU. The images are resized to 640$\times$640. Our model is initialized with MS-COCO \cite{ms_coco} pretrained weights. We decayed the learning rate using the cosine annealing method. 

We evaluate the DEF-YOLO using the traditional object detection metrics. These metrics assess the model’s ability to detect objects across various categories. Key metrics include (a) Mean Average Precision (mAP), which is calculated at Intersection over Union (IoU)=0.5 (mAP@0.5), and (b) across multiple thresholds from 0.5 to 0.95 in steps of 0.05 (mAP@0.5:0.95), providing a comprehensive evaluation of detection performance.  

\subsection{Active THz Dataset for DEF-YOLO Evaluation}
We also report the results on the Active Terahertz (THz) imaging dataset, which consists of 3,157 low-resolution images (5mm$\times$5mm) with 1,194 images containing concealed objects across 11 categories. These objects are hidden on various human body parts, such as the arm, chest, hip, thigh, abdomen, waist, and leg, of 6 male and 4 female subjects standing in either front or back positions. For evaluating DEF-YOLO, we only consider weapon classes (gun, kitchen knife, scissors, metal dagger, ceramic knife, cigarette lighter) to highlight the model’s adaptability in detecting concealed weapons under challenging conditions.

\subsection{Comparison with other Models}
We compare our proposed DEF-YOLO framework against several state-of-the-art object detectors, all pretrained and fine-tuned on TICW and THz datasets. These include RetinaNet\cite{lin2017focal} (2017), which uses FPN and focal loss to manage class imbalance; YOLOv5 \cite{YOLOv5_2020} (2020), which incorporates Mosaic and AutoAugment with CSPDarknet and PANet; and YOLO-X \cite{ge2021YOLOx} (2021), an anchor-free model with dynamic label assignment. Other considered methods for comparison are YOLO-NAS \cite{YOLOnas2021} (2021), using Neural Architecture Search for efficient design; ViTDet \cite{li2022exploring} (2022), adapting Vision Transformers for detection; and YOLOv8 \cite{YOLOv8_2023} (2023), which introduces anchor-free detection and C2f blocks. Also, the recent ones included are Gold-YOLO \cite{wang2023gold} (2023) with self-attention and masked pretraining, YOLOv10 \cite{THU-MIGYOLOv10_2024} (2024) with NMS-free architecture and attention modules, YOLOv11 \cite{YOLO11_ultralytics_2025} (2025) with a transformer backbone and dual label assignment, and YOLO-MS \cite{chen2025YOLO} (2025), which explores multi-scale feature learning through MS-blocks and global query modules.

\section{Results}
\subsection{Quantitative Results}
We start with reporting the best configuration of YOLO that has shown superior performance compared to RetinaNet, ViTDet, and other YOLO versions on our TICW dataset, as shown in the Table \ref{tab:sota}. We observe that YOLOv8 achieves better performance on the TICW dataset for mAP@0.5 (97.8) and mAP@(0.5:0.95) (68.2). We also observe that YOLOv8 adapts better to small-scaled objects, against YOLOv5, due to its anchor-free design that improves the flexibility to object shapes and scales found in thermal images. Furthermore, the lightweight and efficient backbone (C2f modules) improves the detection in YOLOv8 as it results in feature reuse and gradient flow. YOLOv10 and YOLOv11 underperform on thermal images due to their complex architectures, which are less effective for low-detail datasets. In contrast, the simpler and more adaptable YOLOv8 generalizes better on thermal imagery, making it the obvious choice of baseline model for us.

\begin{table}[t]
\begin{center}
\resizebox{0.4\textwidth}{!}{%
\begin{tabular}{|l|cc|cc|}
\hline
Dataset                                                     & \multicolumn{2}{c|}{TICW}                                & \multicolumn{2}{c|}{THz}                                 \\ \hline
 
Method                                                      & \multicolumn{1}{c|}{mAP$_1$} & mAP$_2$ & \multicolumn{1}{c|}{mAP$_1$} & mAP$_2$ \\ \hline
 
RetinaNet \cite{lin2017focal}                                                   & \multicolumn{1}{c|}{84.0} & 56.4 & \multicolumn{1}{c|}{60.0} & 30.6 \\ 
 
YOLOv5 \cite{YOLOv5_2020}                                                    & \multicolumn{1}{c|}{97.7} & 66.7 & \multicolumn{1}{c|}{60.9} & 33.2 \\ 
 
YOLO-X \cite{ge2021YOLOx}                                                      & \multicolumn{1}{c|}{97.0} & 66.0 & \multicolumn{1}{c|}{64.1} & 33.7 \\
 
YOLO-NAS \cite{YOLOnas2021}                                                 & \multicolumn{1}{c|}{61.0} & 43.0 & \multicolumn{1}{c|}{11.1} & 7.10 \\

ViTDet \cite{li2022exploring}                                                 & \multicolumn{1}{c|}{96.9} & 63.8 & \multicolumn{1}{c|}{59.8} & 32.4 \\

YOLOv8 \cite{YOLOv8_2023}                                                  & \multicolumn{1}{c|}{97.8} & 68.2 & \multicolumn{1}{c|}{57.6} & 33.3 \\ 
 
Gold-YOLO \cite{wang2023gold}                                                  & \multicolumn{1}{c|}{96.5} & 66.3 & \multicolumn{1}{c|}{63.9} & 35.4 \\ 
 
YOLOv10 \cite{THU-MIGYOLOv10_2024}                                                    & \multicolumn{1}{c|}{96.7} & 67.2 & \multicolumn{1}{c|}{65.0} & 34.1 \\ 
 
YOLOv11 \cite{YOLO11_ultralytics_2025}                                                    & \multicolumn{1}{c|}{97.5} & 67.8 & \multicolumn{1}{c|}{54.3} & 31.8 \\ 
 
YOLO MS \cite{chen2025YOLO}                                                     & \multicolumn{1}{c|}{95.0} & 60.7 & \multicolumn{1}{c|}{58.6}   & 28.3   \\ 
 
\begin{tabular}[c]{@{}c@{}}DEF-YOLO \\ (ours)\end{tabular} & \multicolumn{1}{c|}{\textbf{98.4}} & \textbf{70.3} & \multicolumn{1}{c|}{\textbf{66.6}} & \textbf{39.4} \\ \hline
\end{tabular}}%
\end{center}
\vspace{-0.1cm}
\captionsetup{font=scriptsize}
\caption{Comparison of detection performance across multiple object detection methods on the TICW and THz dataset. mAP\_1 refer to mAP@0.5 and mAP\_2 to mAP@(0.5:0.95) }
\label{tab:sota}
\end{table}

Next, we present the performance of DEF-YOLO, which surpasses the baseline YOLOv8, thereby demonstrating the effectiveness of our framework. DEF-YOLO achieves the highest mAP@50 of 98.4, surpassing YOLOv8 by +0.6 on the TICW dataset. Moreover, the notable improvement of +2.1 in mAP@0.5:0.95 over YOLOv8 indicates that the proposed model not only detects objects effectively but also achieves superior localization accuracy compared to existing methods. DEF-YOLO enhances YOLOv8’s adaptability and focus by integrating deformable convolution and focal loss, respectively, resulting in a +2.1 improvement in mAP@0.5:0.95 compared to YOLOv8. Additionally, improvements in both mAP metrics on the THz dataset indicate that DEF-YOLO more precisely handles the challenges of low resolution, blurry contours, and low-contrast objects present in THz images, outperforming the listed competing methods.

\subsection{Qualitative Analysis}
\begin{figure*}[t]
\begin{center}
\includegraphics[height=6cm, width=\linewidth]{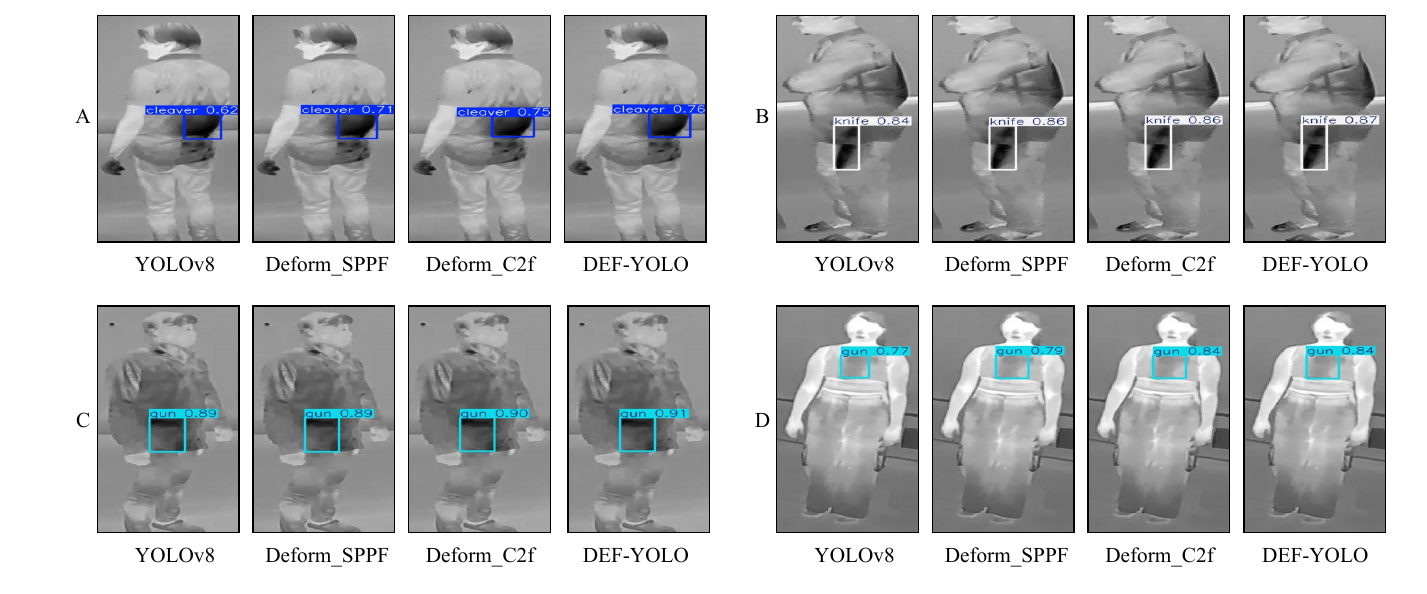}
\end{center}
\vspace{-0.8cm}
\captionsetup{font=scriptsize}
\caption{Examples demonstrating the performance with each modification from YOLOv8 to DEF-YOLO.}
\label{fig:compare}
\end{figure*}

\begin{figure}[t]
\begin{center}
\includegraphics[height=3.5cm, width=\linewidth]{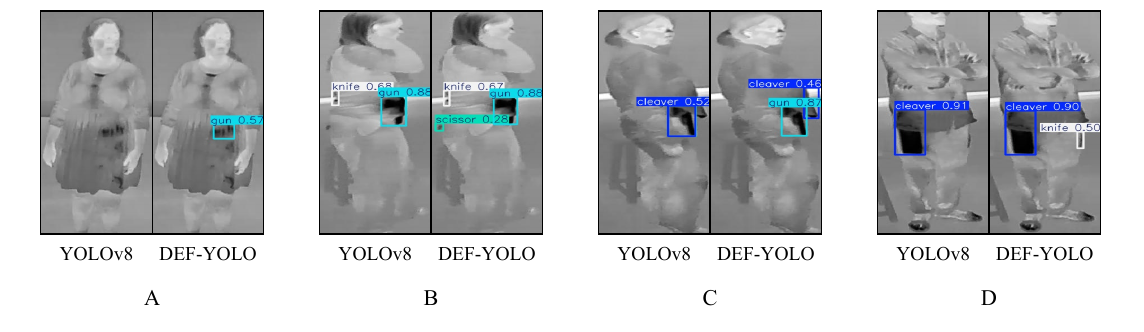}
\end{center}
\vspace{-0.8cm}
\captionsetup{font=scriptsize}
\caption{Examples showing the performance of YOLOv8 and DEF-YOLO.}
\label{fig:base_defYOLO}
\end{figure}

Fig. \ref{fig:compare} shows the progressive improvement achieved through our proposed modifications from YOLOv8 to DEF-YOLO. In A, the detection confidence for the "cleaver" increases remarkably from 0.67 in YOLOv8 to 0.76 in DEF-YOLO, showing enhanced detection of partially occluded (in person's sweat) large-sized objects. In example B, the confidence for "knife" detection increases with each modification, highlighting improved edge and shape representation from Deform\_SPPF and Deform\_C2f. 
In examples C and D, the confidence of "gun" detection improves, especially for cases with background noise or low contrast (D). These observations may conclude that (a) Deform\_SPPF contributes by enabling better spatial aggregation in the presence of irregular objects, (b) Deform\_C2f strengthens the feature extraction pipeline by refining semantics and avoiding loss of feature representations, and finally (c) the DEF-YOLO architecture leverages both enhancements to achieve robust and accurate detection in challenging, real-world thermal imagery. Edge cases such as low visibility, side-view, partial occlusion, and confusing backgrounds are better handled progressively, making DEF-YOLO more reliable for security-critical applications.

Fig. \ref{fig:base_defYOLO} compares the detection results of YOLOv8 and the proposed DEF-YOLO across four challenging scenarios (A–D). In A, DEF-YOLO achieves a score (0.57) in detecting "gun", indicating better sensitivity to low-contrast, partially occluded objects. In contrast, YOLOv8 could not even detect it. In B, YOLOv8 misses a second object entirely ("scissor"), while DEF-YOLO successfully detects it, showing superior multi-object detection in cluttered scenarios. DEF-YOLO detects both "cleaver" and "gun" with improved localization and confidence, demonstrating robustness to overlapping objects and varying poses in example C. Similarly, in D, YOLOv8 misses the "knife", while DEF-YOLO accurately detects both "cleaver" and "knife" with high confidence, reflecting better generalization in detecting small or partially hidden weapons. These results indicate that DEF-YOLO effectively mitigates difficult cases such as object occlusion, multi-class clutter, low contrast, and background confusion, outperforming YOLOv8 in both accuracy and detection completeness.

\begin{table}[t]
\resizebox{\linewidth}{!}{
\centering
\begin{tabular}{|c|ccccc|ccccc|ccccc|ccccc|}
\hline
\textbf{Metric}       & \multicolumn{5}{c|}{\textbf{Precision}}                                                                                                            & \multicolumn{5}{c|}{\textbf{Recall}}                                                                                                               & \multicolumn{5}{c|}{\textbf{mAP@0.5}}                                                                                                              & \multicolumn{5}{c|}{\textbf{mAP@(0.5:0.95)}}                                                                                                       \\ \hline
\textbf{Classes}      & \multicolumn{1}{c|}{\textbf{A}} & \multicolumn{1}{c|}{\textbf{C}} & \multicolumn{1}{c|}{\textbf{G}} & \multicolumn{1}{c|}{\textbf{K}} & \textbf{S} & \multicolumn{1}{c|}{\textbf{A}} & \multicolumn{1}{c|}{\textbf{C}} & \multicolumn{1}{c|}{\textbf{G}} & \multicolumn{1}{c|}{\textbf{K}} & \textbf{S} & \multicolumn{1}{c|}{\textbf{A}} & \multicolumn{1}{c|}{\textbf{C}} & \multicolumn{1}{c|}{\textbf{G}} & \multicolumn{1}{c|}{\textbf{K}} & \textbf{S} & \multicolumn{1}{c|}{\textbf{A}} & \multicolumn{1}{c|}{\textbf{C}} & \multicolumn{1}{c|}{\textbf{G}} & \multicolumn{1}{c|}{\textbf{K}} & \textbf{S} \\ \hline

{YOLOv8}      & \multicolumn{1}{c|}{95.8}       & \multicolumn{1}{c|}{95.6}       & \multicolumn{1}{c|}{96.4}       & \multicolumn{1}{c|}{95.9}       & 95.0       & \multicolumn{1}{c|}{95.1}       & \multicolumn{1}{c|}{95.5}       & \multicolumn{1}{c|}{95.0}       & \multicolumn{1}{c|}{95.4}       & 94.4       & \multicolumn{1}{c|}{97.6}       & \multicolumn{1}{c|}{98.2}       & \multicolumn{1}{c|}{97.7}       & \multicolumn{1}{c|}{97.6}       & 96.8       & \multicolumn{1}{c|}{68.2}       & \multicolumn{1}{c|}{71.6}       & \multicolumn{1}{c|}{73.3}       & \multicolumn{1}{c|}{65.2}       & 64.4       \\ \hline

{Deform\_SPPF} & \multicolumn{1}{c|}{97.1}       & \multicolumn{1}{c|}{94.5}       & \multicolumn{1}{c|}{96.7}       & \multicolumn{1}{c|}{97.2}       & 100.0      & \multicolumn{1}{c|}{93.4}       & \multicolumn{1}{c|}{94.9}       & \multicolumn{1}{c|}{92.7}       & \multicolumn{1}{c|}{94.4}       & 91.8       & \multicolumn{1}{c|}{97.4}       & \multicolumn{1}{c|}{97.3}       & \multicolumn{1}{c|}{97.4}       & \multicolumn{1}{c|}{97.3}       & 97.7       & \multicolumn{1}{c|}{68.6}       & \multicolumn{1}{c|}{71.1}       & \multicolumn{1}{c|}{73.8}       & \multicolumn{1}{c|}{65.5}       & 64.2       \\ \hline

{Deform\_C2f}  & \multicolumn{1}{c|}{96.0}       & \multicolumn{1}{c|}{94.7}       & \multicolumn{1}{c|}{97.6}       & \multicolumn{1}{c|}{96.6}       & 95.2       & \multicolumn{1}{c|}{95.6}       & \multicolumn{1}{c|}{95.5}       & \multicolumn{1}{c|}{95.3}       & \multicolumn{1}{c|}{96.2}       & 95.3       & \multicolumn{1}{c|}{97.6}       & \multicolumn{1}{c|}{97.9}       & \multicolumn{1}{c|}{97.5}       & \multicolumn{1}{c|}{98.0}       & 97.2       & \multicolumn{1}{c|}{69.2}       & \multicolumn{1}{c|}{72.6}       & \multicolumn{1}{c|}{73.7}       & \multicolumn{1}{c|}{65.9}       & 64.5       \\ \hline

{Focal Loss}   & \multicolumn{1}{c|}{96.4}       & \multicolumn{1}{c|}{96.6}       & \multicolumn{1}{c|}{97.8}       & \multicolumn{1}{c|}{96.9}       & 96.4       & \multicolumn{1}{c|}{96.0}       & \multicolumn{1}{c|}{95.7}       & \multicolumn{1}{c|}{95.4}       & \multicolumn{1}{c|}{96.3}       & 95.5       & \multicolumn{1}{c|}{98.4}       & \multicolumn{1}{c|}{98.5}       & \multicolumn{1}{c|}{98.9}       & \multicolumn{1}{c|}{98.2}       & 97.4       & \multicolumn{1}{c|}{70.3}       & \multicolumn{1}{c|}{73.4}       & \multicolumn{1}{c|}{74.1}       & \multicolumn{1}{c|}{66.9}       & 66.0       \\ \hline
\end{tabular}}
\vspace{0.1cm}
\captionsetup{font=scriptsize}
\caption{Ablation Study for the TICW dataset. Where A denotes all classes, C-cleaver, G-gun, K-knife, and S-scissors. }
\label{tab:ablation}
\end{table}

\subsection{Ablation Study}
It is evident from the Table. \ref{tab:ablation} that YOLOv8 struggles in detecting concealed weapons, such as ``knife'' and ``scissors'', achieving a lower mAP@0.5:0.95 of 65.2 and 64.4, respectively. We observe that our first modification (Deform\_SPPF) to baseline YOLOv8 improves the mAP@0.5:0.95 (68.6), precision (97.1) across all classes, with a drop in recall (93.4). The second modification, Deform\_C2f, led to a significant improvement in recall across all classes, compared to the baseline. Also, the mean Average Precision (mAP) at IoU 0.5:0.95 for the ``gun'', ``knife'', and ``scissors'' classes reports the substantial gains over YOLOv8. This suggests that integrating deformable convolutions in the early and mid layers enhances the model’s ability to adapt to occluded and deformed thermal patterns. Furthermore, we observe that the incorporation of focal loss achieves a remarkable boost in the performance, with the highest mAP@0.5:0.95 of 70.3. Focal loss is particularly effective in improving detection of rare classes—such as ``scissors'' and ``cleavers''—in low-contrast thermal images with clutter or occlusion.  
Scissors improved by +1.6 points in mAP@0.5:0.95, which corresponds to a +2.5\% relative improvement over its baseline (64.4). Such relative gains are more pronounced for thin and under-represented objects like scissors and knives than for guns, indicating that focal loss mitigates class imbalance by emphasizing hard samples. Overall, DEF-YOLO demonstrates itself as a more robust and efficient framework for the CWD task in thermal imagery, compared to the baseline model.

\begin{table}[t]
\centering
\resizebox{0.3\linewidth}{!}{
\begin{tabular}{|c|c|c|}
\hline
\textbf{Method}         & \textbf{GFlops} & \textbf{\#params} \\ \hline
YOLOv8                 & 28.7            & 11.137 M              \\ 
Deform\_SPPF & 28.4            & 16.753 M              \\ 
Deform\_C2f & 23.7            & 17.158 M              \\ \hline

\end{tabular}}
\captionsetup{font=scriptsize}
\caption{Effect of model complexity and computational efficiency on various modifications in YOLOv8.}
\label{gflops}
\end{table}

We also present an analysis of model complexity in terms of the number of parameters (in millions, M) and computational efficiency, measured by the number of giga floating-point operations (GFLOPs), for the various modifications applied to YOLOv8 in the development of DEF-YOLO. GFLOPs represent the computational cost required to perform a specific task, as indicated by the number of floating-point operations. As shown in the Table. \ref{gflops}, the baseline YOLOv8 model requires 28.7 GFLOPs and contains 11.137 M parameters. When deformable convolution is introduced into the SPPF layer, there is a slight reduction in GFLOPs (28.4) but a noticeable increase in parameter count (16.753 M). This is because deformable convolutions are heavier in terms of parameter count but may reduce computational redundancy, therefore a slight drop in GFLOPs. The Deform\_C2f configuration further reduces GFLOPs while significantly increasing model complexity. This is due to the extensive use of deformable convolutions in the C2f layers, which substantially raises the parameter count but enhances computational efficiency.

\begin{table}[t]
\centering
\resizebox{\linewidth}{!}{
\begin{tabular}{|c|c|c|c|c|c|c|c|c|c|c|c|}
\hline
\textbf{Method} & RetinaNet & YOLOv5 & YOLO-X & YOLO-NAS & ViTDet & YOLOv8 & Gold-YOLO & YOLOv10 & YOLOv11 & YOLO MS & DEF-YOLO \\ \hline
\textbf{Inf. Time (ms)} & 210 & 2.3 & 19.66 & 3.3 & 91.3 & 1.4 & 1.66 & 1.43 & 1.5 & 6.5 & 3.6 \\ \hline
\textbf{FPS} & 4.76 & 434.78 & 50.86 & 303 & 10.95 & 714 & 602 & 699 & 666 & 153 & 277 \\ \hline
\end{tabular}}
\vspace{0.1cm}
\captionsetup{font=scriptsize}
\caption{Inference time and frame per second of various state-of-the-art methods against our proposed DEF-YOLO.}
\label{tab:fps}
\end{table}

Table. \ref{tab:fps} shows that the proposed method DEF-YOLO achieves a highly competitive balance between inference time (in milliseconds) and frame-per-second (FPS) against several state-of-the-art methods. While methods, such as YOLOv8 and YOLOv10, exhibit extremely low inference times (1.4 ms and 1.43 ms respectively), our proposed framework achieves a strong performance with an inference time of just 3.6 ms, outperforming heavier models, such as YOLO-X (19.66 ms), while still delivering a higher FPS of 277, significantly surpassing models like ViTDet (10.95 FPS) and RetinaNet (4.76 FPS). More importantly, DEF-YOLO strikes an optimal balance between speed and efficiency, making it ideal for real-time, low-latency applications.

\section{Conclusion}
In this paper, we present a novel approach based on YOLOv8 for concealed weapon detection in thermal images. We propose modifications on a few layers of YOLOv8, such as SPPF and bottleneck blocks of C2f layers, to make low, mid, and high-level features adaptive to learn the dynamics of concealed weapons in thermal images, where the objects do not have a definite shape and texture. Another major contribution of the paper is largest comprehensive dataset, TICW, having 6k thermal images with multiple weapons captured from different viewpoints, making it suitable for real-time concealed weapon detection. We also use focal loss along with YOLOv8 loss to handle class imbalance and hard examples. 

Our method achieves the best detection accuracy (98.4\%) and localization precision (70.3\%) and surpasses various competitive object detection models at least by 0.61 and 2.1 in detection accuracy and localization precision, respectively, on the TICW dataset. The proposed model is generic and can be used for images with other modalities. We also tested the proposed model on the Active THz dataset and showed that our model outperforms all the competitive models. The major achievement of the model is to capture the concealed weapons in side-views and in low-contrast thermal images. Although the model improves overall accuracy, deformable convolutions yield class-dependent benefits, and detection of small or thin objects (e.g., knives, scissors) remains limited due to challenges in localizing fine-scale thermal features.

\bibliographystyle{splncs04}
\bibliography{references}

\end{document}